\documentclass[a0,portrait]{a0poster}

\usepackage{multicol} 
\usepackage[svgnames]{xcolor} 

\usepackage{times} 

\usepackage{graphicx} 
\graphicspath{{figures/}} 
\usepackage{booktabs} 
\usepackage[font=small,labelfont=bf]{caption} 
\usepackage{amsfonts, amsmath, amsthm, amssymb} 
\usepackage{wrapfig} 

\begin{document}



\begin{minipage}[b]{0.75\linewidth}
\VeryHuge \color{NavyBlue} \textbf{Deep frame interpolation} \color{Black}\\ 
\Huge\textit{Frame interpolation using deep neural networks with optical flow as a prior}\\[2.4cm] 
\huge \textbf{Vladislav Samsonov}\\[0.5cm] 
\huge Moscow Institute of Physics and Technology, Department of Innovation and High Technology\\[0.4cm] 
\Large \texttt{vvladxx@gmail.com} --- +7 (985) 270 8365\\
\end{minipage}
\begin{minipage}[b]{0.25\linewidth}
\includegraphics[width=11cm]{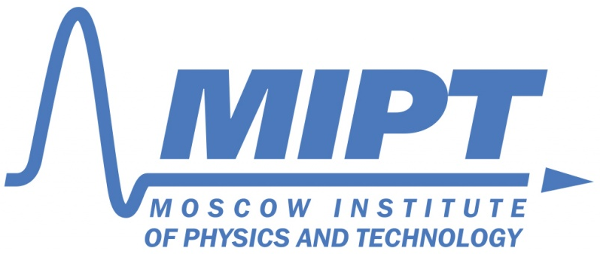}\
\includegraphics[width=7cm]{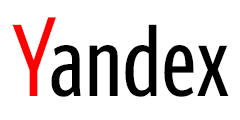}\\
\end{minipage}

\vspace{1cm} 


\begin{multicols}{3} 


\color{Navy} 

\begin{abstract}
This work presents a supervised learning based approach to the computer vision problem of frame interpolation. The presented technique could also be used in the cartoon animations since drawing each individual frame consumes a noticeable amount of time. The most existing solutions to this problem use unsupervised methods and focus only on real life videos with already high frame rate. However, the experiments show that such methods do not work as well when the frame rate becomes low and object displacements between frames becomes large. This is due to the fact that interpolation of the large displacement motion requires knowledge of the motion structure thus the simple techniques such as frame averaging start to fail. In this work the deep convolutional neural network is used to solve the frame interpolation problem. In addition, it is shown that incorporating the prior information such as optical flow improves the interpolation quality significantly.
\end{abstract}

\color{Black}
\section*{Introduction}
Frame interpolation is one of the most challenging tasks in the computer vision. The goal of the frame interpolation is to increase the number of frames in a video sequence to make it more visually appealing. Numerous approaches were proposed to solve this problem \cite{JiefuZhai} \cite{SimoneMeyer}. However, these approaches are unsupervised and do not exploit particular video structure. This work presents a deep convolutional neural network for frame interpolation. Deep convolutional neural networks are known to be one of the best methods in machine learning to extract semantic information. Frames with a small displacement typically do not require high-level semantic information. But when the motion becomes complex and the object displacements between consecutive frames becomes large, this high-level information becomes crucial to effectively restore the middle frame. Deep convolutional networks were successfully applied for similar tasks before. The G. Long et al. \cite{DBLP:journals/corr/LongKAL16} proposed a deep convolutional method for computing the frame interpolation in order to obtain the optical flow. The P. Fischer et al. \cite{DBLP:journals/corr/FischerDIHHGSCB15} trained deep convolutional network for direct computation of optical flow. Compared to these methods the method proposed in this paper solves the problem of frame interpolation instead of the optical flow thus the visual quality of the middle frame is of high importance. The B. Yahia \cite{BENYAHIA} used convolutional network to generate the middle frame but the result was blurry and visually unpleasant. In this paper it is shown how to overcome this problem by moving from the MSE loss objective to the adversarial training. It is also shown how to further improve quality by incorporating the optical flow prior.

\color{Black} 

\section*{Network architecture}
\begin{center}\vspace{1cm}
\includegraphics[width=0.33\linewidth]{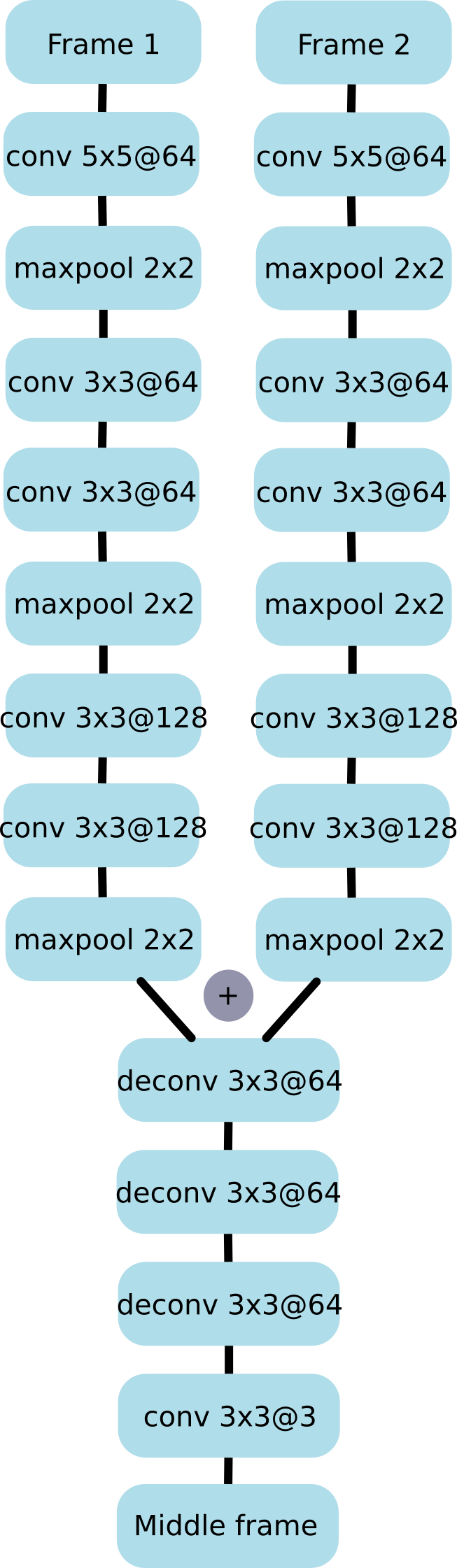}
\captionof{figure}{Network architecture used for frame interpolation}
\end{center}
The Y-style neural network with separate inputs for the first and for the second frame is used. The weights between the first and second line are shared to enforce symmetric output. As the side-effect this trick effectively reduces dimensionality and prevents overfitting as well. Two lines are then merged by elementwise sum and upsampled again using transposed convolution (also called deconvolution sometimes). Following \cite{DBLP:journals/corr/HeZRS15} the residual connections between corresponding downsampling and upsampling layers are added but this is not shown on the illustration.

\section*{Evaluation}
The two network models are trained: one with simple mean-squared-error (MSE) and one with adversarial approach suggested by the work of I. Goodfellow et al. \cite{DBLP:journals/corr/GoodfellowPMXWOCB14}. It is shown that adversarial approach helps to overcome unnatural blurriness of the output image.\\
For the dataset the open source movie Sintel was used. It is also one of the standard datasets for the optical flow evaluation. Since the existing MPI Sintel dataset is tailored mainly for the optical flow and too small to train deep convolutional network, we splitted the full movie into the sequence of 21312 frames. Then we took every consecutive triplets of frames for the training samples. The triplet consists of the first frame, ground truth middle frame and the second frame.
\subsection*{MSE loss objective}
First the network was trained with a simple MSE loss as in \cite{BENYAHIA} \cite{DBLP:journals/corr/LongKAL16}:
$$L(I_1, I_2) = \sum_{i=1}^{n}\sum_{j=1}^{m} (I_1^{ij} - I_2^{ij})^2$$
\begin{center}\vspace{1cm}
\includegraphics[width=0.3\linewidth]{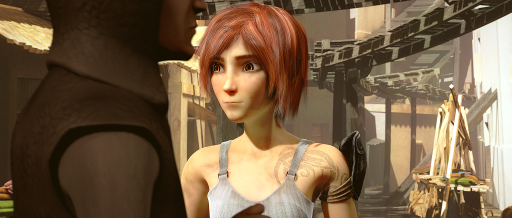}
\includegraphics[width=0.3\linewidth]{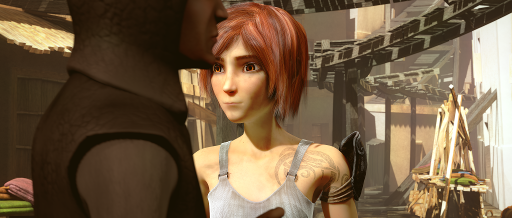}
\includegraphics[width=0.3\linewidth]{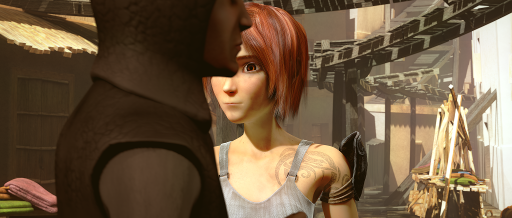}
\includegraphics[width=0.3\linewidth]{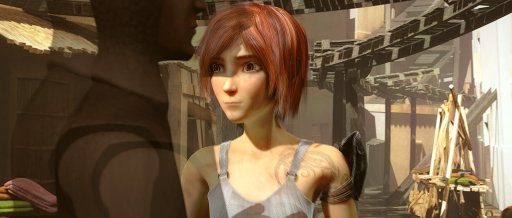}
\includegraphics[width=0.3\linewidth]{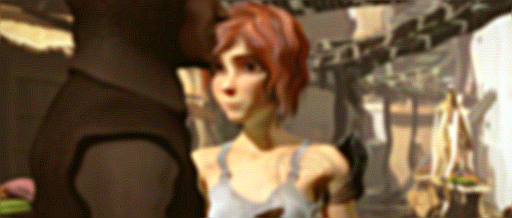}
\includegraphics[width=0.3\linewidth]{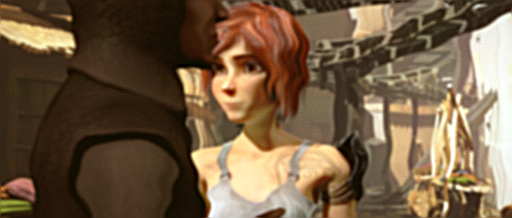}
\captionof{figure}{From left to right, from top to bottom: first frame, ground truth frame, second frame, average of the first and second, output of the network with the MSE loss, output of the adversarial network}
\end{center}
As it can be seen from the example output, MSE loss leads to unnatural blurriness of the output image.

\subsection*{Adversarial approach}
Several methods were proposed to construct the loss function which is close to the human perception such as \cite{DBLP:journals/corr/JohnsonAL16}. We will use the adversarial training approach proposed by I. Goodfellow et al. \cite{DBLP:journals/corr/GoodfellowPMXWOCB14}. The adversarial approach jointly trains two networks: generator network and discriminator network. The first one generates the middle frame while the second one outputs the probability that this frame is generated by the first network and is not drawn from the original distribution. The second network is used as the loss function by the first one. The first network tries to minimize this probability while the second tries to distinguish generated frames from the original frames. LeNet-like architecture with 16 layers was used for the discriminator classifier network. To increase convergence speed the MSE loss was used for the initial training steps with an exponential decay:

$$L(I_1, I_2) = \alpha \sum_{i=1}^{n}\sum_{j=1}^{m} (I_1^{ij} - I_2^{ij})^2 + D(I_1, I_2)$$

Here $D$ denotes discriminator network and $\alpha$ descreasing with an exponential decay:

$$\alpha = e^{-\gamma n}$$

The result of the adversarial approach looks more visually appealing compared to the previous method yet shows slightly worse values according to MSE and PSNR.

\section*{Incorporating optical flow prior}
Computation of the optical flow has been studied extensively in the computer vision. Various methods were proposed for this task, but we choose DeepFlow for it's state-of-the-art performance and available open-source implementation.\\
Formally, optical flow is a vector field where each vector component shows relative displacement of a point.\\
To incorporate optical flow prior to the existing architecture, we introduce a new layer type called Displacement Convolutional Layer (DCL). It is the generalization of the regular convolutional layer:
$$C_{ij} = \sum_{k=-w}^{w} \sum_{l=-w}^{w} I(i+d_y(i,j)+k, j+d_x(i,j)+l) W(k,l)$$
where $d_x(i,j)$ and $d_y(i,j)$ are the $x$ and $y$ components of the optical flow vector in position $(i,j)$. If we set $d \equiv 0$, we will get the regular convolutional layer as a special case of DCL:
$$C_{ij} = \sum_{k=-w}^{w} \sum_{l=-w}^{w} I(i+k, j+l) W(k,l)$$

\begin{center}
\includegraphics[width=0.6\linewidth]{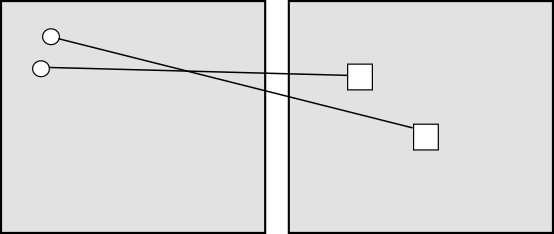}
\captionof{figure}{Illustration of the Displacement Layer}
\end{center}
Finally, the only thing needed to incorporate the optical flow prior is to replace the first convolutional layer by the Displacement Layer. Our final network with optical flow prior is trained using adversarial approach as before.
\begin{center}\vspace{1cm}
\includegraphics[width=0.45\linewidth]{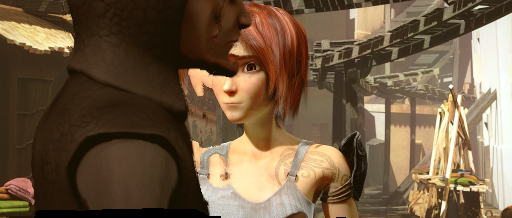}
\includegraphics[width=0.45\linewidth]{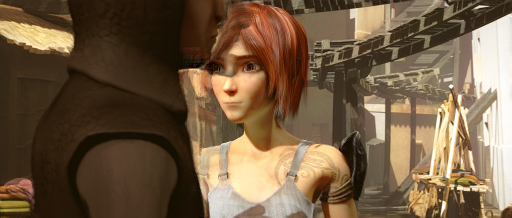}
\captionof{figure}{From left to right: simple warping from the optical flow, output of the network with the optical flow prior}
\end{center}
Given the perfect optical flow, the middle frame could be reconstructed by the simple warping. Thus, the presented approach is also compared with the simple warping from the optical flow field to validate the necessity of neural network and demonstrate that the neural network learns to compensate for the optical flow inaccuracy.
\begin{center}
\includegraphics[width=0.75\linewidth]{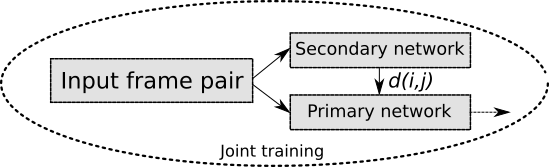}
\end{center}
Note that in this case optical flow is needed for training along with the video frames. Calculating accurate optical flow for the frame sequence might be a problem. It is actually possible to get rid of this requirement. To achieve this we introduce another neural network which is trained to predict vector field $d(i,j)$ of DCL. Both networks are trained jointly as a whole with the same loss function as before. This network learns to predict optical flow implicitly from the frame sequence which means that frame sequence alone is enough for training: there is no need to calculate optical flow beforehand.

\begin{center}\vspace{1cm}
\begin{tabular}{l l l l}
\toprule
\textbf{} & \textbf{MSE} & \textbf{PSNR} & \textbf{SSIM} \\
\midrule
Average frame & 0.0079 & 21.0 & 0.836 \\
NN with MSE loss & 0.0050 & 23.0 & 0.614 \\
Adversarial NN & 0.0053 & 22.8 & 0.721 \\
Simple warping & 0.0052 & 22.8 & 0.907 \\
NN with optical flow prior & \textbf{0.0023} & \textbf{26.4} & \textbf{0.945} \\
\bottomrule
\end{tabular}
\captionof{table}{Comparison of all methods}
\end{center}\vspace{1cm}


\nocite{*} 
\bibliography{references}

\begin{thebibliography}{1}

\bibitem{JiefuZhai}
J.~Zhai, K.~Yu, J.~Li, and S.~Li, ``A low complexity motion compensated frame
  interpolation method,'' {\em SPIE}, 2005.

\bibitem{SimoneMeyer}
S.~Meyer, O.~Wang, H.~Zimmer, M.~Grosse, and A.~Sorkine-Hornung, ``Phase-based
  frame interpolation for video,'' {\em CVPR}, 2015.

\bibitem{DBLP:journals/corr/LongKAL16}
G.~Long, L.~Kneip, J.~M. Alvarez, and H.~Li, ``Learning image matching by
  simply watching video,'' {\em CoRR}, vol.~abs/1603.06041, 2016.

\bibitem{DBLP:journals/corr/FischerDIHHGSCB15}
P.~Fischer, A.~Dosovitskiy, E.~Ilg, P.~H{\"{a}}usser, C.~Hazirbas, V.~Golkov,
  P.~van~der Smagt, D.~Cremers, and T.~Brox, ``Flownet: Learning optical flow
  with convolutional networks,'' {\em CoRR}, vol.~abs/1504.06852, 2015.

\bibitem{BENYAHIA}
H.~B. Yahia, ``Frame interpolation using convolutional neural networks on 2d
  animation,'' 2016.

\bibitem{DBLP:journals/corr/HeZRS15}
K.~He, X.~Zhang, S.~Ren, and J.~Sun, ``Deep residual learning for image
  recognition,'' {\em CoRR}, vol.~abs/1512.03385, 2015.

\bibitem{DBLP:journals/corr/GoodfellowPMXWOCB14}
I.~J. Goodfellow, J.~Pouget{-}Abadie, M.~Mirza, B.~Xu, D.~Warde{-}Farley,
  S.~Ozair, A.~C. Courville, and Y.~Bengio, ``Generative adversarial
  networks,'' {\em CoRR}, vol.~abs/1406.2661, 2014.

\bibitem{DBLP:journals/corr/JohnsonAL16}
J.~Johnson, A.~Alahi, and F.~Li, ``Perceptual losses for real-time style
  transfer and super-resolution,'' {\em CoRR}, vol.~abs/1603.08155, 2016.

\end{thebibliography}
\bibliographystyle{ieeetr}


\end{multicols}
\end{document}